\documentclass[twoside]{article}
\usepackage{graphicx} 
\usepackage{subfigure} 
\usepackage{natbib}

\usepackage{url}  

\usepackage{algorithm}
\usepackage{algorithmic}

\usepackage{epstopdf}
\usepackage{times,bm}
\usepackage{algorithm}
\usepackage{algorithmic}[1]

\usepackage{subfigure,amssymb,mathrsfs,amsthm,amsfonts,amsmath}
\usepackage{amsmath,graphicx,caption,color,epsfig,float,pbox,tabularx,wrapfig,mathrsfs,multirow,array,diagbox}

\usepackage{mathdots}
\usepackage{amsmath,graphicx,caption,color,epsfig,float,pbox,tabularx,wrapfig,mathrsfs,multirow,amsfonts,amsthm,times,subfigure,amssymb,mathrsfs,algorithm}
\usepackage{epstopdf}

\newcommand{\RB}{\mathbb{R}}
\newcommand{\EB}{\mathbb{E}}

\newtheorem{lemma}{Lemma}
\newtheorem{theorem}{Theorem}

\begin{document}

\title{Accelerated Markov Chain Monte Carlo Using Adaptive Weighting Scheme}

\author{Yanbo Wang$^{1}$ Wenyu Chen$^{1}$ and Shimin Shan$^{3}$}
\date{%
    $^1$ School of Computer Science and Technology, North University of China, Taiyuan\\
    $^2$ School of Semiconductor and Physics, North University of China, Taiyuan\\
}
\maketitle 

\begin{abstract}
Gibbs sampling is one of the most commonly used Markov Chain Monte Carlo (MCMC) algorithms due to its simplicity and efficiency. It
cycles through the latent variables, sampling each one from its distribution conditional on the current values of all the other variables.
Conventional Gibbs sampling is based on the systematic scan (with a deterministic order of variables).
In contrast, in recent years, Gibbs sampling with random scan has shown its advantage in some scenarios. 
However, almost all the analyses of Gibbs sampling with the random scan are based on uniform selection of variables.
In this paper, we focus on a random scan Gibbs sampling method that selects each latent variable non-uniformly.
Firstly, we show that this non-uniform scan Gibbs sampling leaves the target posterior distribution invariant.
Then we explore how to determine the selection probability for latent variables. 
In particular, we construct an objective as a function of the selection probability and solve the constrained optimization problem.
We further derive an analytic solution of the selection probability, which can be estimated easily. 
Our algorithm relies on the simple intuition that choosing the variable updates according to their marginal probabilities enhances the mixing time of the Markov chain.
Finally, we validate the effectiveness of the proposed Gibbs sampler by conducting a set of experiments on real-world applications.
\end{abstract}

\section{Introduction}
\label{sec:introduction}
Gibbs sampling~\citep{geman1984stochastic} is one of the most popular methods for approximate inference, which is required for many models nowadays. Gibbs sampling gained its popularity thanks to its simplicity, efficiency and theoretical convergence guarantee. It is the foundation of multiple probabilistic programming tools such as the automatic statistician \citep{ghahramani2015probabilistic}  and the Anglican \citep{tolpin2016design}.
Moreover, it is widely used in image processing~\citep{geman1984stochastic}, natural language processing~\citep{griffiths2004finding,heinrich2005parameter,yue2024biomamba}, bioinformatics~\citep{liu1994collapsed,pinoli2014latent,zhang2021ddn2,fu2024ddn3,lu2024drugclip,du2024abds} and even deep learning~\citep{lecun2015deep,wang2024twin,lu2018multi,lu2019integrated,xu2024smiles}. Due to the popular use of Gibbs sampling, further advancement of Gibbs sampling is beneficial for a large range of machine learning applications. 

In this paper, we focus on the scan order of Gibbs sampling. Accelerated Gibbs sampling is needed to adapt to big data and complex models nowadays. Recent advancement in computational efficiency for sampling methods includes parallel and distributed computing \citep{angelino2016patterns, magnusson2017sparse},  hybrid sampling method \citep{mimno2012sparse, hoffman2017learning, li2017approximate} and convergence acceleration ~\citep{chen2016herded,he2016scan,de2016ensuring,chen2021data}. Techniques developed for these directions can be further combined to achieve better performance. Research has shown that various scan orders of Gibbs sampling lead to different convergence speeds \citep{he2016scan}. Hence, in this paper, we propose a weighted scan order for Gibbs sampling named weighted Gibbs sampling.

Conventional Gibbs sampling is based on \emph{systematic scan} (also known as the deterministic or sequential scan), which cycles through the latent variables, sampling each one from its distribution conditioned on the current values of all other variables, as shown in Algorithm~\ref{alg:systemscangibbs}.
In the systematic scan, a fixed permutation is selected, and the variables are repeatedly selected in that order. 

In recent years, Gibbs sampling with random scan has been explored. In this method, the variable to sample is selected uniformly and independently at random at each iteration.
It has shown its advantages of mixing time in certain scenarios~\citep{roberts2015surprising,lu2021cot,lu2024uncertainty,lu2023machine,lu2019integrated,he2016scan,de2016ensuring}.
It is shown in Algorithm~\ref{alg:randomscangibbs}.

\begin{algorithm}[t]
	\caption{Systematic Scan Gibbs Sampler}
	\label{alg:systemscangibbs}
	\begin{algorithmic}[1]
		\REQUIRE $x \in \RB^d$ is the parameter of interest, starting position $x^{0}\in \RB^d$, target distribution $\pi(x)$ and $d$ conditional distributions $\pi_1(x_1 \vert x_{-1}), \cdots, \pi_d(x_d \vert x_{-d})$, where $x_{-i} = [x_1,\cdots,x_{i-1},x_{i+1},\cdots,x_{d}]$.
		\FOR {$t = 1, 2, \ldots$}
		\STATE $i \equiv (t - 1 )(\text{mod}\  d) + 1$. \# select the variable to be sampled
		\STATE $x^{t}_i \sim \pi_i(\ \cdot\  \vert\ x^{t-1}_{{-i}})$. \# sample the variable from the distribution conditioned on other variables
		\STATE $x^t = [x_i^t\ ,\ x_{-i}^{t-1}]$. \# only update the $i$-th variable
		\ENDFOR	
	\end{algorithmic}
\end{algorithm}

However, sampling all the variables uniformly may not always be the optimal choice. 
For example, imagine a multivariate Gaussian distribution, where the marginal variances between latent variables are distributed inhomogeneously. 
If we could assign more weight to sample the elongated valley, the sample could traverse the parameter space more efficiently, and we could obtain a more ``fair'' evaluation of the target distribution. We demonstrate this in detail with experiments in Section~\ref{sec:Toyexperiment}.




We thus attempt to devise a non-uniform sampling scan order. We assign each variable a different weight, which is obtained to optimize the effective sample size and select the variable based on this weight.
The main contribution of this paper can be summarized as:
\begin{itemize}
	\item We provide theoretical analysis that a non-uniform Gibbs sampler would still converge to the target distribution.
	\item We derive the selection probability for a non-uniform scan Gibbs sampler.
	This probability is optimal under certain assumptions.
	More importantly, it has a simple form and can be estimated easily.
	\item The proposed method is evaluated in three different experimental settings. Empirical results show that the weighted Gibbs sampling method converges faster than traditional methods, especially where the marginal variances between latent variables are distributed inhomogeneously. 
\end{itemize}

The remainder of the paper is organized as follows: Section~\ref{sec:related} briefly introduces related work.
In Section~\ref{sec:formulation}, we formulate the problem and derive the selection probability. 
Empirical results are reported in Section~\ref{sec:experiment}
and Section~\ref{sec:conclusion} is devoted to the conclusion and future work.
Detailed proof of some theoretical results and additional empirical results are provided in the supplementary materials.
\begin{algorithm}[t]
	\caption{Random Scan Gibbs Sampler~\citep{he2016scan}}
	\label{alg:randomscangibbs}
	\begin{algorithmic}[1]
		\REQUIRE $x \in \RB^d$ is the parameter of interest, starting position $x^{0}\in \RB^d$, target distribution $\pi(x)$ and $d$ conditional distributions $\pi_1(x_1 \vert x_{-1}), \cdots, \pi_d(x_d \vert x_{-d})$. Selection probability $q_1=\cdots=q_d = \frac{1}{d}$.
		\FOR {$t = 1, 2, \ldots$}
		\STATE Select $i$ from $\{1,\ldots,d\}$ with probability $q_i$. 
		\STATE $x^{t}_i \sim \pi_i(\ \cdot\  \vert\ x^{t-1}_{{-i}})$.
		\STATE $x^t = [x_i^t\ ,\ x_{-i}^{t-1}]$.
		\ENDFOR	
	\end{algorithmic}
\end{algorithm}

\section{Related Work}
\label{sec:related}

We propose to use a nonuniform sampling of the variables for Gibbs sampling to accelerate the convergence. We thus briefly review the related work in Gibbs sampling with accelerated convergence and nonuniform sampling in optimization.

\paragraph{Gibbs Sampling with Accelerated Convergence} In recent years, Gibbs sampling has drawn lots of attention owing to its efficiency and simplicity.
For example, \cite{chen2016herded} proposed the herded version of Gibbs sampling.
\cite{tripuraneni2015particle} applies the particle version to infinite Hidden Markov Model.
\cite{de2015rapidly} studied its usage on some factor graphs. 
\cite{de2016ensuring} discussed the asynchronous version and make it scalable to big data.
\cite{he2016scan} explored random-scan Gibbs sampler in general cases and derived some theoretical results about mixing time.

\paragraph{Non-uniform Sampling in Optimization}
Simultaneously, other machine learning communities have focused on weighted sampling in recent years.
For example, \cite{perekrestenko2017faster} argued that applying weighted sampling to each coordinate would accelerate the procedure of coordinate descent.
In the context of Stochastic Gradient Descent (SGD) and Stochastic Variational Inference (SVI)~\citep{hoffman2013stochastic}, weighted sampling has also been proven to outperform the uniform one~\citep{zhao2015stochastic,zhangdeterminantal}.

Specifically, coordinate descent shares some similarities with Gibbs sampling. 
In the optimization community, coordinate descent using random sweep and weighted sweep were proposed one after another and achieve state-of-the-art performance in coordinate descent methods~\citep{liu2015asynchronous,liu2015asynchronous2,perekrestenko2017faster}.

However, non-uniform sampling of the variables for random scan Gibbs sampling has never been explored. In this paper, we fill in the blank and show that non-uniform sampling can be used to accelerate Gibbs sampling.

\section{Weighted Gibbs Sampler}
\label{sec:formulation}

We propose a weighted Gibbs sampler. It is 
a random scan using the Gibbs sampling method, but each latent variable is selected non-uniformly. We first show that such a method with a non-uniform sampling of the variable is unbiased and converges to the same target distribution in Section~\ref{sec:non_uniform}. Then, we introduce the objective of our method, which is to maximize the effective sample size, in Section~\ref{sec:objective}. Based on this formulation, we derive the analytic solutions of the weights for selecting the latent variables and summarize the algorithm in Section~\ref{sec:solution}.

\subsection{Non-uniform Scan Gibbs Sampling}
\label{sec:non_uniform}
First, we prove that non-uniform scan Gibbs sampling leaves the target posterior distribution invariant. This is the foundation of such non-uniform sampling methods. 
\begin{theorem}
	\label{thm:invariant}
	The stationary distribution of non-uniform scans is the expected target distribution.  
\end{theorem}

\paragraph{Proof Sketch}
Specifically, we consider a Markov chain on state space $\Omega$ with transition matrix $P$ and stationary distribution $\pi$, following \cite{he2016scan}.
Let $P_i$ be the transition matrix corresponding to sampling variable $x_i$. 
Also, we use $q_i$ ($i\in \{1,\cdots,d\}$) to represent the sampling probability of the $i$-th variable and we have $\sum_{i=1}^{d} q_i = 1$.

For ease of exposition, we augment the state space.
The augmented state space is $\Psi  = \Omega \times [n]$, including both the current state and the index of the variable to be sampled.
The transition probability is $P((x,i),(y,j)) = q_j P_i(x,y)$.
	
Then we show that applying the transition matrix for non-uniform scan does not change the expected target distribution $\pi $, which satisfies that $\pi ((x,i)) = q_i \pi(x)$.
That is,
\begin{equation*}
	\begin{aligned}
\sum_{x,i}^{} \pi((x,i)) P((x,i),(y,j))
=
q_j \cdot \pi(y) =  \pi (y,j)\\
	\end{aligned}
\end{equation*}
Thus, the stationary distribution of non-uniform scan is $\pi ((x,i)) = q_i \pi(x)$.
Proved. 

\subsection{Maximize the Effective Sample Size}
\label{sec:objective}

Now we focus on finding a good selection probability.
To do this, we formulate the problem as maximizing the Effective Sample Size(ESS)~\citep{robert2004monte} with respect to sampling probability $q = [q_1, \cdots, q_d ]$.
Then we seek a solution to $q$ that optimizes the objective function (Section~\ref{sec:solution}). 


In MCMC-based sampling methods, we want to draw more ``independent'' samples from the desired posterior distribution and reduce the mixing time of the Markov chain.
However, the mixing time is generally intractable.
To describe the independence of draws, a common measurement is the Effective Sample Size (ESS)~\citep{robert2004monte}.
The ESS of a parameter sampled from an MCMC is the number of effectively independent draws from the posterior distribution that the Markov chain is equivalent to.
It is defined as
\begin{equation}
\label{eqn:ess}
\begin{aligned}
\text{ESS} = \frac{N}{1 + 2 \sum_{k=1}^{\infty} \rho_k},
\end{aligned}	
\end{equation}
where $N$ is the number of samples, $\rho_k$ is the autocorrelation of the sampler with lag $k$.
$\big(1 + 2 \sum_{k=1}^{\infty} \rho_k\big)^{-1}$ is also known as the asymptotic efficiency of an MCMC sampler. 
Our goal is to maximize ESS given a fixed $N$, equivalent to minimizing $\sum_{k=1}^{\infty} \rho_k$.

Despite its appeal, this measure is problematic because the higher order autocorrelations are hard to estimate. 
To circumvent this problem, \citet{pasarica2010adaptively,wang2013adaptive} resorted to the Expected Squared Jumping Distance (ESJD), defined as:
\begin{equation}
\label{eqn:esjd}
\begin{aligned}
\text{ESJD}(q) = \EB_q \Vert x^{t+1} - x^{t} \Vert^2.
\end{aligned}	
\end{equation}
In this paper, $\Vert z \Vert$ represents the $l_2$ norm of the vector $z$. 
It has been proven that maximizing the ESJD governed in Equation~\eqref{eqn:esjd} is equivalent to minimizing the first-order autocorrelation $\rho_1$~\citep{pasarica2010adaptively}. 
These results can be easily generalized to any lag $k$.
That is, maximizing $\EB \Vert x^{t+k} - x^{t} \Vert^2$ is equivalent to minimizing $\rho_k$.

Therefore, rather than directly minimizing $\sum_{k=1}^{\infty} \rho_k$, we choose to solve the following optimization problem as a surrogate: 
\begin{equation}
\label{eqn:optimize}
\begin{aligned}
& \underset{q=\{q_1,\cdots,q_d\}}{\arg\max}\  \sum_{k=1}^{\infty} \EB_{q} \Vert x^{t+k} - x^{t} \Vert^2\\
& \text{such that } \sum_{i=1}^{d} q_i = 1, \ q_i > 0 .
\end{aligned}	
\end{equation}

\subsection{Analytic Form of the Sampling Weights}
\label{sec:solution}

To solve the optimization problem (Equation~\eqref{eqn:optimize}), we attempt to expand $ \EB_{q} \Vert x^{t+k} - x^{t} \Vert^2$ into a more tractable form. 
To proceed, we introduce an assumption.
Temporarily, we ignore the correlation between different dimensions.
That is, here we restrict our interest to sampling from the posterior distribution of the following form 
\begin{equation}
\begin{aligned}
\pi(x) = \prod_{i=1}^d \pi_i(x_i),
\end{aligned}	
\end{equation}
where $\pi_i(x_i)$ is the marginal distribution for variable $x_i$.
We call it \emph{mean-field approximation}.
Then we decompose $k$-th order ESJD as
\begin{equation}
\begin{aligned}
\EB_{q} \Vert x^{t+k} - x^{t} \Vert^2 = \sum_{i=1}^{d} \EB_{q} (x_i^{t+k} - x_i^{t})^2.
\end{aligned}	
\end{equation}

Owing to the  \emph{mean-field approximation}, we have $\pi_i(x_i \vert x_{-i}) = \pi_i(x_i)$. 
That is to say, sampling the $i$-th variable is equivalent to independently sampling $x_i$ from its marginal distribution $\pi_i(x_i)$.
Thus, once the $i$-th sampling variable is visited/updated between $t$-th and $t+k$-th iterate, 
$x_i^{t+k} - x_i^{t}$ is irrelavant to the times that $i$-th sampling variable is visited during the period.
If the $i$-th variable is not updated during the period, then $x_i^{t+k} - x_i^{t} = 0$.

Thus, when we analyze $x_i^{t+k} - x_i^{t}$, we can divide it into two classes: ``visited''  and ``unvisited'' and expand $\EB_{q} \Vert x^{t+k} - x^{t} \Vert^2$ as:
\begin{equation}
\label{eqn:visit}
\begin{aligned}
& \EB_{q} \Vert x^{t+k} - x^{t} \Vert^2
=  \sum_{i=1}^{d} \EB_{q} (x_i^{t+k} - x_i^{t})^2\\
= & \sum_{i=1}^{d} \big[\text{Pr}\{\text{i is visited}\}\cdot d_i + \text{Pr}\{\text{i is unvisited}\}\cdot 0 \big] \\
= & \sum_{i=1}^{d} \text{Pr}\{\text{i is visited}\}\cdot d_i,\\
\end{aligned}	
\end{equation}
where $d_i$ is defined as
\begin{equation}
\label{eqn:d}
\begin{aligned}
{d}_i &=  \iint (x_i^{t+k} - x_i^{t})^2 \pi_i(x_i^{t+k}) \pi_i(x_i^{t}) dx_i^{t+k} dx_i^{t}\\
& = 2 \EB_{\pi_i}(x_i^2) - 2(\EB_{\pi_i}(x_i))^2 = 2 \text{Var}_{\pi_i}(x_i).
\end{aligned}	
\end{equation}
Once $x_i$ is updated, $x_i^t$ and $x_i^{t+k}$ can be seen as two independent samples from $x_i$'s marginal distribution according to the mean-field assumption.
We also have
\begin{equation}
\begin{aligned}
\text{Pr}\{\text{i is visited}\} = & 1 - \text{Pr}\{\text{i is unvisited}\}\\
= &  1 - (1-q_i)^k. \\
\end{aligned}
\end{equation}

Then Equation~\eqref{eqn:visit} can be rewritten into 
\begin{equation}
\label{eqn:visit2}
\begin{aligned}
\EB_{q} \Vert x^{t+k} - x^{t} \Vert^2 = \sum_{i=1}^{d} \big[\big(  1 - (1-q_i)^k\big)\cdot d_i \big]\\
\end{aligned}	
\end{equation}

Problem~\eqref{eqn:optimize} can be simplified as
\begin{equation}
\label{eqn:optimize3}
\begin{aligned}
& \underset{\{q_1,\cdots,q_d\}}{\arg\max}\  \sum_{k=1}^{\infty} \sum_{i=1}^{d} \big[\big(  1 - (1-q_i)^k\big)\cdot d_i \big],\\
& \text{such that } \sum_{i=1}^{d} q_i = 1, \ q_i > 0.
\end{aligned}	
\end{equation}

Dropping terms of the above objective function that do not vary with $q_i$, we are left with the optimization problem
\begin{equation}
\label{eqn:optimize5}
\begin{aligned}
& \underset{\{q_1,\cdots,q_d\}}{\arg\min}\  \sum_{k=1}^{\infty} \sum_{i=1}^{d} \big[ (1-q_i)^k \cdot d_i \big],\\
& \text{such that } \sum_{i=1}^{d} q_i = 1, \ q_i > 0 .
\end{aligned}	
\end{equation}

Then we solve this constrained optimization problem. 
\begin{lemma}
	\label{lem:solution}
	The solution to optimization problem~\eqref{eqn:optimize5} is 
	\begin{equation}
	\label{eqn:solution}
	\begin{aligned}
	q_i = \frac{\sqrt{{d_i}} }{\sum_{j=1}^{d}\sqrt{{d_j}} } \ \ \ \text{for } i = 1,\cdots,d . \\
	\end{aligned}	
	\end{equation}
\end{lemma}

\paragraph{Proof Sketch}	
First, we rearrange the objective function $\sum_{k=1}^{\infty} \sum_{i=1}^{d} \big[ (1-q_i)^k \cdot d_i \big]$ into a much more convenient form:
\begin{equation}
\label{eqn:optimize4}
\begin{aligned}
 \sum_{k=1}^{\infty} \sum_{i=1}^{d} \big[ (1-q_i)^k \cdot d_i \big]
	=  \sum_{i=1}^{d}  \frac{1}{q_i} {d}_i. \\
	\end{aligned}	
	\end{equation}
	
Then, combining the constraint that $\sum_{i=1}^{d} q_i = 1, \ q_i > 0$, we apply Cauchy-Schwarz inequality to obtain the lower bound of the objective function:
\begin{equation}
\label{eqn:optimize6}
\begin{aligned}
	& \sum_{k=1}^{\infty} \sum_{i=1}^{d} \big[ (1-q_i)^k \cdot d_i \big] 
	=  \sum_{i=1}^{d}  \frac{1}{q_i} {d}_i \\
\geq & \big(\sum_{i=1}^d \sqrt{\frac{1}{q_i} {d}_i  q_i} \big)^2 = \big(\sum_{i=1}^d \sqrt{ {d}_i} \big)^2.\\ 
	\end{aligned}	
	\end{equation}
	
The equality is obtained only when $\frac{1}{q_1^2} {d}_1 = \frac{1}{q_2^2} \hat{d}_2  = \cdots = \frac{1}{q_d^2} {d}_d $.
Combined with the constraint that $\sum_{i=1}^{d} q_i = 1, \ q_i > 0$, we have that 
\begin{equation}
\begin{aligned}
q_i = \frac{\sqrt{{d_i}} }{\sum_{j=1}^{d}\sqrt{{d_j}} } \ \ \ \text{for } i = 1,\cdots,d .
\end{aligned}	
\end{equation}
	Thus, we conclude the proof. 
	
\begin{algorithm}[t]
	\caption{weighted Gibbs sampler}
	\label{alg:main}
	\begin{algorithmic}[1]
		\REQUIRE $x$ is the parameter of interest, starting position $x^{0}\in \RB^d$, integer $k>1$, target distribution $\pi(x)$ and $d$ conditional distributions $\pi_1(x_1 \vert x_{-1}), \cdots, \pi_d(x_d \vert x_{-d})$.  $q_1,\ldots,q_d = \frac{1}{d}$.
		\FOR {$t = 1, 2, \ldots$}
		\STATE Select $i$ from $\{1,\ldots,d\}$ with probability $q_i$. 
		\STATE $x^{t}_i \sim \pi_i(\ \cdot\  \vert\ x^{t-1}_{{-i}})$.
		\STATE $x^t = [x_i^t\ ,\ x_{-i}^{t-1}]$.
		\STATE Update $q_1,\ldots,q_d$ every $k$ iterations via Equation~\eqref{eqn:regularization}.
		\ENDFOR	
	\end{algorithmic}
\end{algorithm}

\paragraph{Intuition}
The selection probability described in Equation~\eqref{eqn:solution} is consistent with the intuition that choosing the variable updates according to their marginal probabilities can make the samples more ``independent''.
For those variables that have large (marginal) variance, we tend to sample these variables more frequently to have a ``fair'' evaluation of the posterior distribution.

\paragraph{Estimate $\hat{d}_i$} 
In the general case, owing to the complexity of the posterior, $\pi_i(x_i)$ is intractable, so is $d_i$. 
Thus, we choose to use $\hat{d}_i$ instead of $d_i$. According to Equation~\eqref{eqn:d}, ${d}_i$ is proportional to the variance of $\pi_i(x_i)$, so it can be estimated empirically. 
The resulting algorithm is listed in Algorithm~\ref{alg:main}.
In practice, to obtain the weight for every variable, we sequentially scan all the variables several times (usually 1-3) before weighted sampling.
Since the sampling procedure is intrinsically stochastic, for any $i \in \{1,\cdots,d\}$, $\hat{d}_i$ may come arbitrarily close to $0$. 
This motivates introducing a small positive regularization parameter $\lambda$ to bias the estimate of $\hat{d}_i$. 
In such cases, sampling probability becomes
\begin{equation}
\label{eqn:regularization}
\begin{aligned}
q_i = \frac{\sqrt{\hat{d_i}} + \lambda }{\sum_{j=1}^{d}\big( \sqrt{\hat{d_j}} + \lambda \big) }  \ \ \ \text{\ for } i = 1,\cdots,d . 
\end{aligned}	
\end{equation}

\section{Experiments}
\label{sec:experiment}

In this section, we conduct applications of our method with empirical evaluations. To demonstrate the behavior of the method, we first evaluate it with a Gaussian distribution using synthetic data in various settings (Section~\ref{sec:Toyexperiment}). Then, we carry out two different experiments with real-world applications. First, we apply weighted Gibbs sampling to Markov Random Field (MRF)~\citep{li2009markov,chen2024trialbench} for an image denoising task~\citep{lu2023deep,yi2018enhance}. After that, we evaluate our method using Latent Dirichlet Allocation (LDA)~\citep{blei2003latent} to learn topics from text data.
We employ both systematic scan Gibbs sampler and random scan Gibbs sampler as baseline methods.
All three sampling methods share the same setup and initialization for fair comparison. 

\subsection{Gaussian Distribution with Synthetic Data}
\label{sec:Toyexperiment}

We conduct an experiment on a high-dimensional Gaussian distribution to demonstrate the behavior of different methods under various conditions. This is the same example as discussed in the introduction Section~\ref{sec:introduction} (a high-dimensional Gaussian distribution, and the variances of different variables differs greatly).

We use two different target distributions to illustrate the performance of different methods. In the first case, the target distribution is comparably homogeneous. While in the second case, it is relatively heterogeneous.  We expect that higher weights would be assigned to the sampling variables with high variances, according to Equation~\eqref{eqn:solution}.
We use $\Sigma = \lambda I + \epsilon YY^\top \in \RB^{d\times d}$ to set the covariance matrix to generate synthetic data, where $I\in \RB^{d\times d}$ is the identity matrix, $Y \in \RB^{d\times r}$ is generated by random sampling from a normal distribution. 
$d$ is fixed to 50 in this task.
Via adjusting $r$, $\lambda$ and $\epsilon$, we can control the covariance matrix.

The first Gaussian distribution is a homogeneous one, each variable has the same order of variance, where we let $r = 50$, $\epsilon = 0.1$ and $\lambda = 10$. 
Correspondingly, the second distribution is heterogeneous, where we set $r = 5$, $\epsilon = 5$ and $\lambda = 10$.
We draw $2\times 10^4$ samples for all methods and tasks.
The first $2\times10^3$ samples are regarded as burn-in samples.

\begin{figure}[t]	
\centering
	\subfigure[Homogeneous]{
		\centering
		\includegraphics[width=3.7cm]{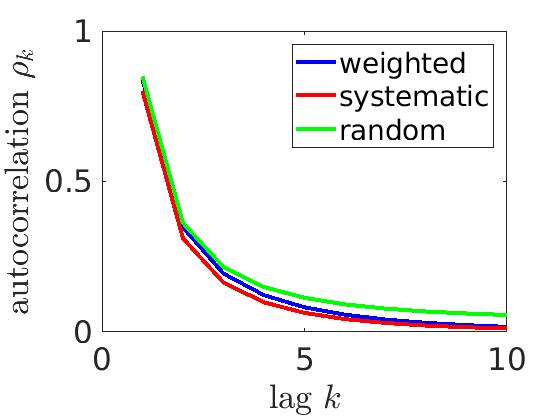}		\label{fig:toy_1}}
        \hspace{-10pt}
	\subfigure[Heterogeneous]{
		\centering
		\includegraphics[width=3.7cm]{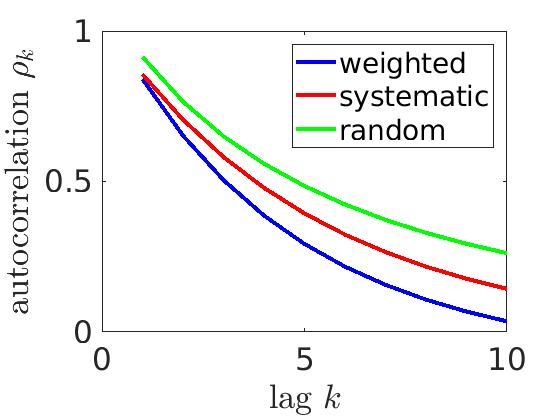}
		\label{fig:toy_2}}
\caption{Synthetic data experiment on a Gaussian distribution. 
For the homogeneous case, the gaps between all three methods are small.
The weighted Gibbs sampler outperforms baseline methods clearly in the heterogeneous case.}
	\label{fig:toy}
\end{figure}

\begin{figure*}[p]	
\centering
\subfigure[Random scan]{
		\centering
		\includegraphics[width=4.7cm, height = 3cm]{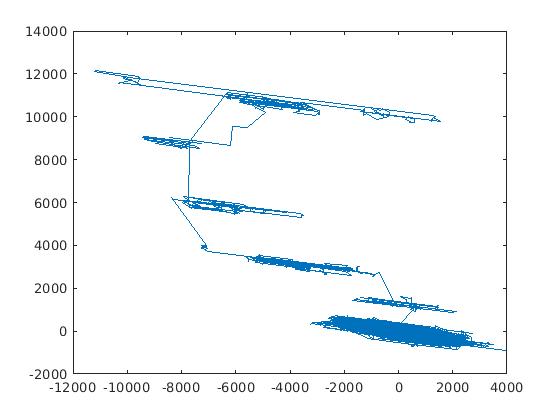}
		\label{fig:toy_3}}
        \hspace{-10pt}
	\subfigure[Systematic scan]{
		\centering
		\includegraphics[width=4.7cm, height = 3cm]{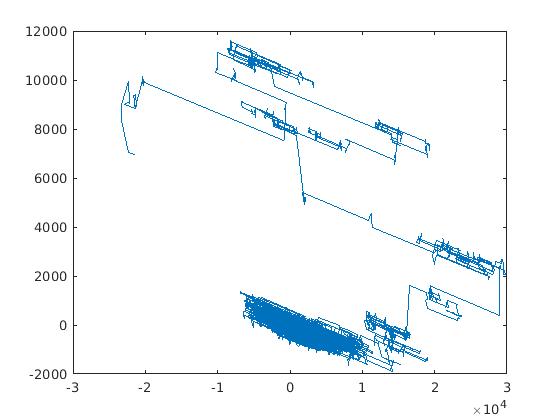}
		\label{fig:toy_4}}
\subfigure[Our method]{
		\centering
		\includegraphics[width=4.7cm, height = 3cm]{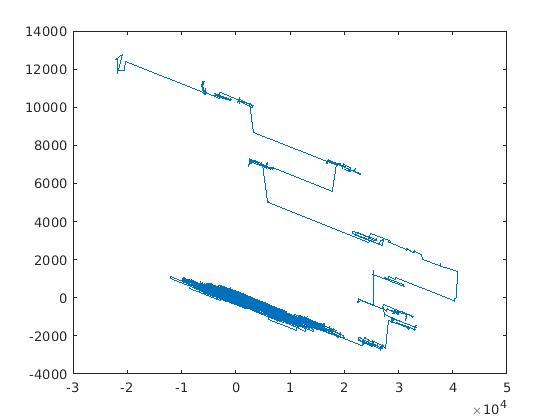}
		\label{fig:toy_5}}  
        \vspace{-5pt}
\caption{Visualization of sampling steps using different sampling methods from the synthetic data experiment. Since the experiment is highly dimensional, the first two principal components are employed for this visualization using PCA. It shows that the burn-in period is shorter using the proposed weighted Gibbs sampling. }    
	\label{fig:toy2}
\end{figure*}

To measure the correlation between samples, we estimate autocorrelation coefficients~\citep{robert2004monte,wu2022cosbin}.
The results for the first and second distributions are shown in Figure~\ref{fig:toy_1} and~\ref{fig:toy_2}, respectively.
We find that for the homogeneous case, systematic scan performs best among all the methods with a very small margin.
In contrast, our method clearly outperforms baseline methods for the heterogeneous distribution.
We also explore the burn-in period of all three methods.
Concretely, we perform principal component analysis (PCA) on all the samples and reduce the dimension to 2 for ease of visualization.
We draw the trace of first $2\times 10^3$ samples in Figure~\ref{fig:toy2}. 
We find that for the weighted Gibbs sampler it takes a short time to reach the high-probability region, thus it can reduce the burn-in time.


\begin{figure*}[p]
\centering
	\subfigure[Clean binary image]{
		\centering
		\includegraphics[width=4.4cm]{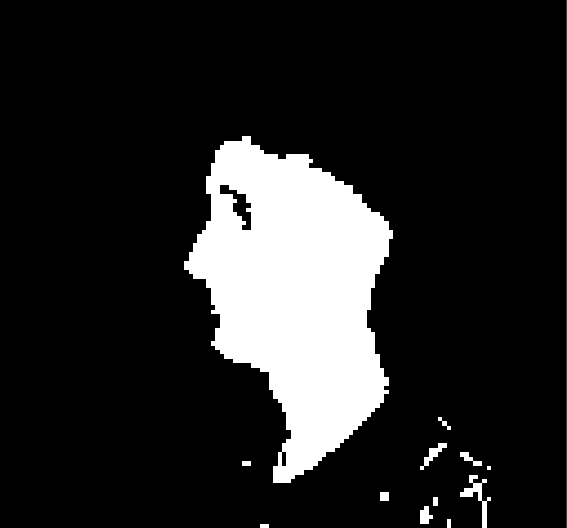}
		\label{fig:bo_origin}}
	\subfigure[Noisy image]{
		\centering
		\includegraphics[width=4.4cm]{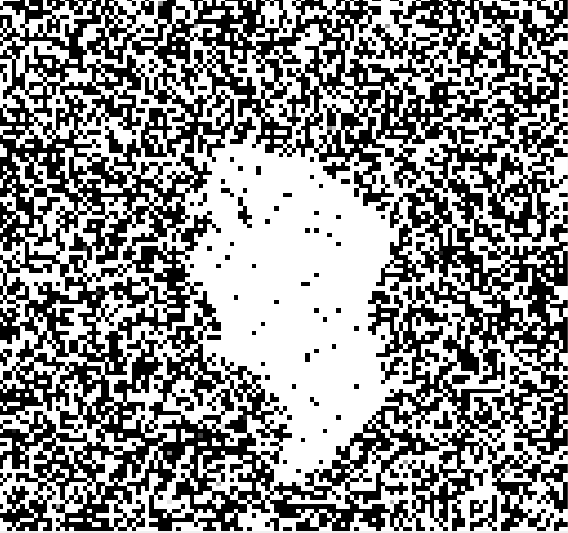}
		\label{fig:bo_noise}}
	\subfigure[Recovered image]{
		\centering
		\includegraphics[width=4.4cm]{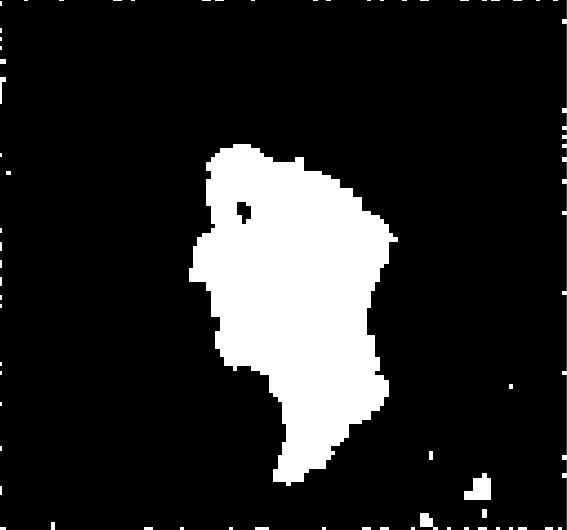}
		\label{fig:bo_recover}}		
        \vspace{-5pt}
	\caption{An exemplar case of image denoising using Grid-Lattice Markov Random Field.  In this example, the noise variance is set to $\sigma = 1.0$.}	
	\label{fig:bo1}
\end{figure*}

\begin{figure*}[p]
	\subfigure[$\sigma= 0.7$]{
		\centering
		\includegraphics[width=5.7cm,height =4cm]{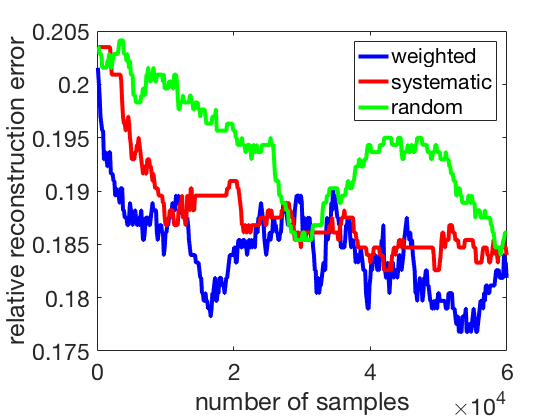}
		\label{fig:bo_a}}
	\subfigure[$\sigma = 0.8$]{
		\centering
		\includegraphics[width=5.7cm,height =4cm]{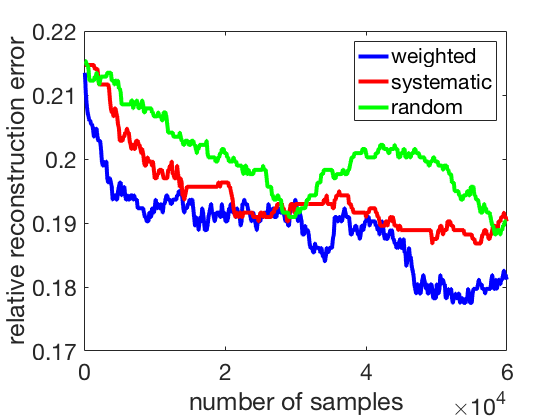}
		\label{fig:bo_b}}	
	\subfigure[$\sigma = 0.9$]{
		\centering
		\includegraphics[width=5.7cm,height =4cm]{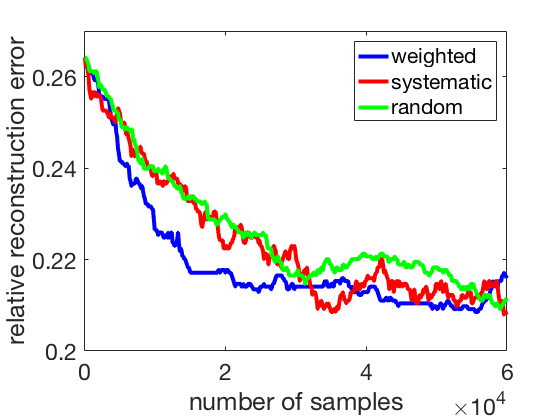}
		\label{fig:bo_c}}
	\subfigure[$\sigma = 1.0$]{
		\centering
		\includegraphics[width=5.7cm,height =4cm]{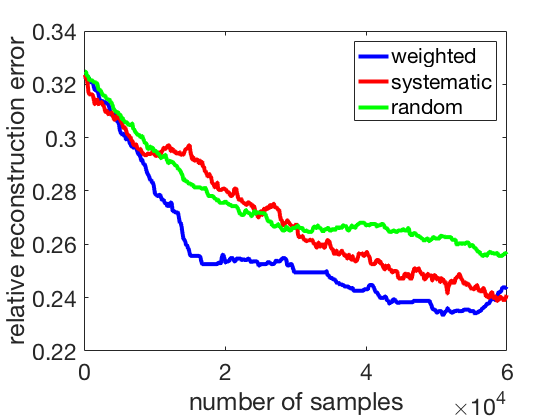}
		\label{fig:bo_d}}
	\subfigure[weight distribution for various variances ($\sigma = 0.5,0.7,1$, respectively)]{
		\centering
		\includegraphics[width=5.7cm,height =4cm]{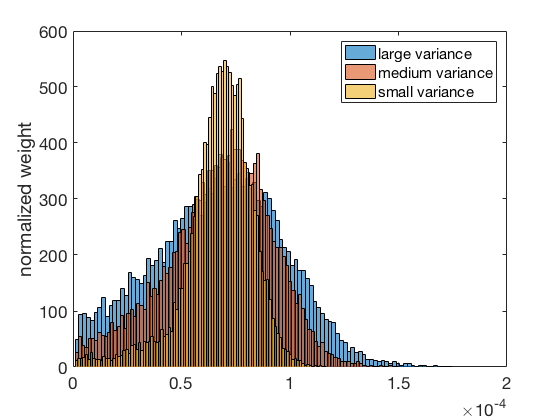}
		\label{fig:bo_e}}
	\subfigure[weight of different pixels for $\sigma = 1.0$]{
		\centering
		\includegraphics[width=5.7cm,height =4cm]{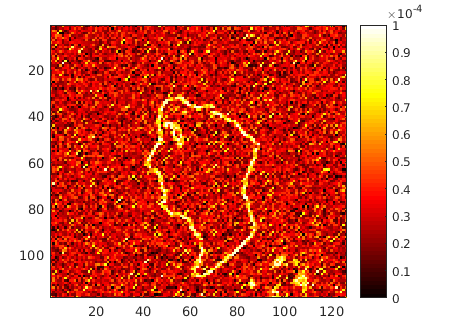}
		\label{fig:bo_f}}		
	\caption{Evaluation of the image denoising task using Grid-Lattice Markov Random Field. 
Panel (a)-(d) show relative $L_2$ reconstruction errors as a function of the number of draws for different variances. 
Then we display the weight distribution for different $\sigma$ in panel (e). 
In panel (f), the weights of different pixels are reported.
We also conduct the same procedure to other three images in the dataset and report the results in supplementary materials.  
}	
\label{fig:bo2}
\end{figure*}

\begin{figure}[]	
\centering
\centering
\subfigure[Graphical representation of 8 topics.]{
\includegraphics[width=7.2cm]{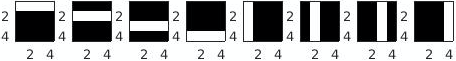}	
\label{fig:syn_topic}}
\subfigure[Examplar documents.]{
\includegraphics[width=7.2cm]{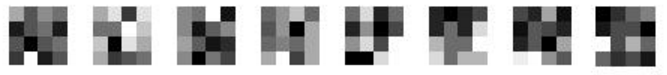}
\label{fig:documents}}
 \caption{Synthetic data examples. Panel (b) shows a subset of the generated images (documents), where each image is the results of 100 samples from a unique mixture of these topics shown in panel (a).}	
\label{fig:lda_synData}
\end{figure}

\begin{figure}[]
\centering
\includegraphics[width=6.7cm]{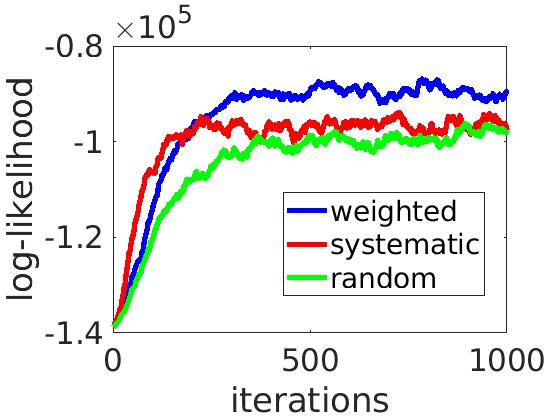}
 \caption{The change of log-likelihood over iterations using topic model with synthetic data. It is worth mentioning that each iteration means the algorithm passes through $N$ data points (Suppose the training set has $N$ data points). For systematic scan Gibbs sampling, each variable is updated exactly once during one iteration. 
We find that the weighted Gibbs sampler converges faster than the baseline methods.}	
\label{fig:lda_syn_logLikelihood}
\end{figure}


\begin{figure}[]
\centering
\includegraphics[width=6.0cm]{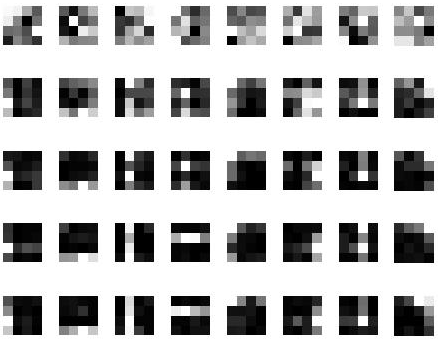}	
 \caption{Estimation of the topics over time using synthetic data.
 Each line represents the learned topics at certain iterations.
 They correspond to 1, 50, 100, 500, 1000 iterations.
 The learned topic gradually approaches the true topic.}	
\label{fig:lda_topic}
\end{figure}


\begin{figure}[]
\centering
\subfigure[R8]{
\centering
\hspace{-10pt}
\includegraphics[width=4.0cm]{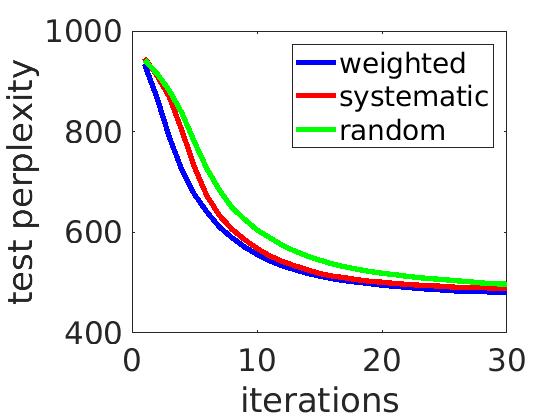}
\label{fig:lda_a}}
\hspace{-10pt}
\subfigure[R52]{
\centering
\includegraphics[width=4.0cm]{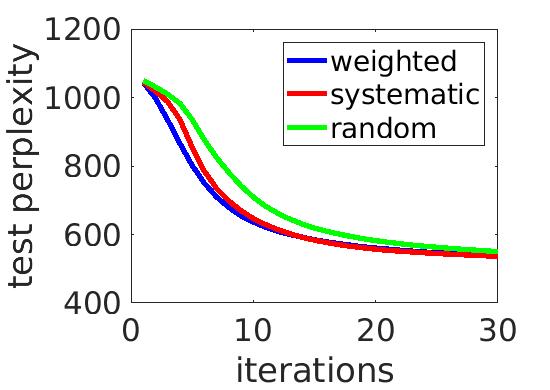}
\label{fig:lda_b}}		
\caption{Test perplexity over sampling iterations using topic Model with R8 and R52 datasets.
Lower perplexity indicates better performance. 
For all three methods, we first sequentially scan all the sampling variables and then perform different sampling strategies.}	
\label{fig:lda_1}
\end{figure}



\begin{figure}[]	
\centering
\hspace{-10pt}
	\subfigure[R8]{
		\centering
		\includegraphics[width=4.0cm]{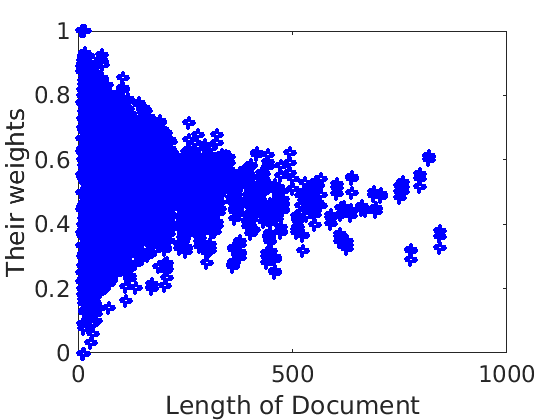}
		\label{fig:lda_c}}
        \hspace{-10pt}
	\subfigure[R52]{
		\centering
		\includegraphics[width=4.0cm]{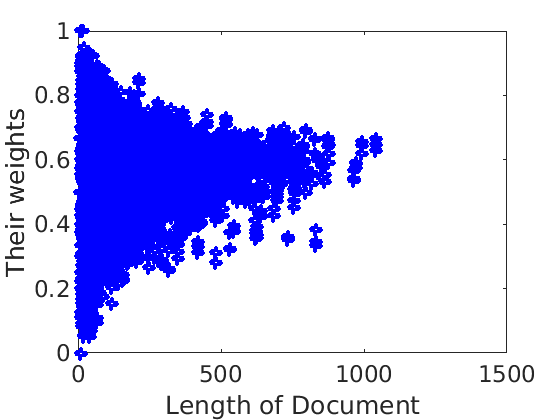}
		\label{fig:lda_d}}
        \vspace{-5pt}
	\caption{The relationship between the length of documents and their weights using the topic Model with R8 and R52 datasets. }	
	\label{fig:lda_2}
\end{figure}

\subsection{Grid-lattice MRF for Image Denoising}
\label{sec:MRFexperiment}

We now evaluate our method on a grid-lattice Markov Random Field (MRF) for an image denoising task~\citep{chen2016herded,lu2022cot,xu2024mambacapsule}. 
Suppose that we are given a noise-corrupted image $Y$ and we want to recover the original binary image $X$.
$x_i \in \{-1, +1\}$ denotes the true value of the $i$-th pixel, which is unknown. 
The variable $y_i$ represents the observed value of the $i$-th pixel, i.e., the noise-corrupted value of this pixel. 

We use the CMU Face Images dataset\footnote{\url{https://archive.ics.uci.edu/ml/datasets.html}}.
We select one representative image from the dataset.
Three other images are processed using the same procedure, and the results are presented in the supplementary materials.
First, we turn it into a binary image, shown in Figure~\ref{fig:bo_origin}.
Noisy versions of the binary image ( Figure~\ref{fig:bo_noise}) are created through the addition of Gaussian noise with zero mean and $\sigma^2$ variance.

To make use of the fact that neighboring pixels are likely to have the same label, we impose an Ising prior.
Specifically, the prior for $X$ is
$p(X) \propto \exp\big(\frac{1}{2} \sum_{i \sim j}J_{ij}x_i x_j\big)$,
where $i\sim j$ indicates that nodes $i$ and $j$ are connected. 
$J_{ij}$ describes the coupling strength between nodes $i$ and $j$.
For simplicity, we assume all the coupling parameters $J_{ij}$ are identical, i.e., $J_{ij} = J$. 
Furthermore, it is assumed that the likelihood terms $p(y_i \vert x_i)$ are conditionally independent (a Gaussian with known variance $\sigma^2$ and mean $x_i$). 
Thus the posterior of $X$ is 
\begin{equation}
\begin{aligned}
& p(X \vert Y, J, \sigma )  \propto p(X) \prod_i p(y_i \vert x_i) \\
= & \exp\big(\frac{1}{2} \sum_{i \sim j} J x_i x_j - \frac{1}{2\sigma^2} \sum_{i } (y_i - x_i)^2\big).
\end{aligned}
\end{equation}

We compare the weighted Gibbs sampler against both systematic and random scan Gibbs samplers.
We measure the relative $L_2$ reconstruction errors, which is defined as $\frac{\Vert X_{\text{true}} - \hat{X} \Vert_{F}}{\Vert X_{\text{true}}\Vert_{F}}$, where $\hat{X}$ is the recovered value.
The plot of the reconstruction errors with respect to the number of draws using different methods are shown in Figure~\ref{fig:bo_a},~\ref{fig:bo_b},~\ref{fig:bo_c} and ~\ref{fig:bo_d}.  For all these methods, we sequentially scan all the sampling variables during the first iteration.
In this period, all three plots overlap, thus the plot starts from the second iteration. 
It shows that the weighted Gibbs sampler converges to a lower error faster than two baseline methods in most cases, especially when the variances of the added noise are large.

We also show the weight distribution for different noise in Figure~\ref{fig:bo_e}. With different levels of noise in the images, the weights are in general non-uniform distributed. 
We find that a larger variance of the noise  would cause a less uniform distribution of weight, where our method usually works better.
Also, we show the variance of different pixels in Figure~\ref{fig:bo_f}, from which we observe that more weights are assigned to the boundary region.
This means our method would sample these variables more frequently and thus decrease the reconstruction error more rapidly.
An example image recovered by the weighted Gibbs sampler is shown in Figure~\ref{fig:bo_recover}. 
We can see that the noise is greatly removed, compared with the noise-corrupted image in Figure~\ref{fig:bo_noise}.
The results of other images in the dataset are shown in supplementary materials.

\subsection{Topic Model with Latent Dirichlet Allocation}
\label{sec:LDAexperiment}
Finally, we apply our method to Latent Dirichlet Allocation (LDA)~\citep{blei2003latent}, which is a topic model. 
Gibbs sampling has been a popular inference method for topic models \cite{griffiths2004finding,heinrich2005parameter}.

We first demonstrate the behavior of our method using a synthetic dataset which is easy to visualize. Then, we evaluate it with Reuters news datasets for news topic representation learning. 
It is worth mentioning that our weighted sampling is based on each document. 
That is, we assign each document a weight.
Thus, we only require storing additional vectors (whose length is a number of documents), and it doesn't have a high demand for memory. 

\paragraph{A Graphical example} Firstly, following \cite{griffiths2004finding}, we generate a small dataset in which the output of the algorithm can be shown graphically. 
The dataset contains 2,000 images. 
Each of them has 16 pixels in a $4 \times 4$ grid, where each pixel corresponds to a word. 
The intensity of any pixel is specified by a nonnegative integer, which represents the word count.
This dataset is of exactly the same form as a word document cooccurrence matrix constructed from a database of documents, with each image being a document, with each pixel being a word, and with the intensity of a pixel being its frequency.
The images were generated by defining a set of 8 topics corresponding to horizontal and vertical bars shown in Figure~\ref{fig:syn_topic} and exemplar documents are shown in Figure~\ref{fig:documents}.
We set the document length to be 100. 
We use open-source code\footnote{\url{http://xiaodong-yu.blogspot.com/2009/08/matlab-demo-of-gibbs-sampling-for.html}} for this task.

Figure~\ref{fig:lda_syn_logLikelihood} shows the log-likelihood over sampling iterations comparing different sampling methods. We can see that our proposed method obtained the highest log-likelihood with a short burn-in time. Additionally, we plot the per-topic word distribution at different iterations in  Figure~\ref{fig:lda_topic}. The learned topics (Figure~\ref{fig:lda_topic}) appear very similar to the ground-truth (Figure~\ref{fig:syn_topic}).  

\paragraph{R8 and R52 datasets} The datasets we use are R8 and R52 datasets from Reuters 21578, which are available online\footnote{\url{ http://csmining.org/index.php/r52-and-r8-of-reuters-21578.html}}.
Reuters-21578 R8 dataset contains 11976 words in vocabulary, 5485 documents in the training set and 2189 documents in the testing set.
Reuters-21578 R52 dataset contains 22292 words in the vocabulary, 6532 documents in the training set and 2568 documents in the testing set. 
For both datasets, we use 100 topics.   We adapt our method to an open-source implementation
\footnote{Code is available at:  \url{http://gibbslda.sourceforge.net/}}.

Perplexity on the test dataset with respect to Gibbs sampling iteration is shown in Figure~\ref{fig:lda_1}.
We observe that our method can converge to a lower perplexity than the other two baseline methods within the first few iterations. 
Furthermore, we explore the relationship between the length of documents and their corresponding weights.
Weights are drawn from three randomly chosen iterations (not burn-in).
The results are shown in Figure~\ref{fig:lda_2}, and we find that longer documents usually produce stable weights, while for shorter documents, their weights differ greatly.

\section{Conclusion and Future Work}
\label{sec:conclusion}
In this paper, we have proposed a novel weighted Gibbs sampler. We proved that non-uniform sampling methods can be used for random scan Gibbs sampling while keeping the same stationary distribution.
In particular, we cast the original problem of finding proper weights into an optimization problem of maximizing the effective sample sizes. Moreover, we obtained an analytic solution of the optimal weights, which can be estimated efficiently. Thus, our method can be easily implemented and easily adopted in any existing implementation. The proposed method demonstrates the advantages of mixing time through various experimental settings. 

We plan to continue the future work in two directions.
Firstly, a natural idea is to extend the proposed method into an asynchronous setting.
It can increase the scalability of the proposed methods.
Moreover, we will pursue a theoretical guarantee on the accelerated mixing time of the proposed weight Gibbs sampler in the future.

\bibliographystyle{uai}
\bibliography{ref}


\end{document}